\documentclass[letterpaper, 10 pt, conference]{ieeeconf}  

\IEEEoverridecommandlockouts                              

\usepackage{color}

\usepackage{graphics} 
\usepackage{amsmath,amssymb,amsfonts} 
\usepackage{bm}
\usepackage{subcaption}	 
\usepackage{graphicx}
\usepackage{multirow}
\usepackage{array}

\usepackage{url}
\usepackage{hyperref}
\hypersetup{colorlinks,urlcolor=red}
\usepackage{xcolor}

\usepackage{amsmath}

\usepackage{booktabs}
\usepackage{multirow}
\usepackage{textcomp}
\usepackage{setspace}

\usepackage{pifont}
\newcommand{\cmark}{\ding{51}}
\newcommand{\xmark}{\ding{55}}

\newcommand{\nt}[1]{\textcolor[rgb]{0.0,0.0,1.0}{\textbf{#1}}}

\newcommand{\bP}{\mathbf{P}}

\newcommand{\bT}{\mathbf{T}}
\newcommand{\bp}{\mathbf{p}}

\newcommand{\ba}{\mathbf{a}}

\newcommand{\br}{\mathbf{r}}

\newcommand{\bg}{\mathbf{g}}

\newcommand{\bb}{\mathbf{b}}

\newcommand{\bR}{\mathbf{R}}

\newcommand{\bv}{\mathbf{v}}

\newcommand{\be}{\mathbf{e}}

\newcommand{\bPhi}{\boldsymbol{\Phi}}

\usepackage[backend=bibtex,natbib=true,style=numeric-comp,sorting=none,giveninits=true,maxbibnames=99,url=false,doi=false]{biblatex}
\title{\LARGE \bf
Ctrl-VIO: Continuous-Time Visual-Inertial Odometry \\ for Rolling Shutter Cameras
}

\author{Xiaolei Lang$^{1,\dagger}$, Jiajun Lv$^{1,\dagger}$, Jianxin Huang$^1$, Yukai Ma$^1$, Yong Liu$^{1,*}$, Xingxing Zuo$^{2,*}$
\thanks{ $^1$ The authors are with the Institute of Cyber-Systems and Control, Zhejiang University, Hangzhou, China. }%
\thanks{$^2$ The author is with the Department of Informatics,
Technical University of Munich, Germany.}
\thanks{$^\dag$The co-first authors have equal contributions.}
\thanks{$^*$ Yong Liu and Xingxing Zuo are the corresponding authors (Email: {\tt\small yongliu@iipc.zju.edu.cn; xingxing.zuo@tum.de}).}
}

\begin{document}

\maketitle
\thispagestyle{empty}
\pagestyle{empty}

\begin{abstract}
In this paper, we propose a probabilistic continuous-time visual-inertial odometry (VIO) for rolling shutter cameras. The continuous-time trajectory formulation naturally facilitates the fusion of asynchronized high-frequency IMU data and motion-distorted rolling shutter images. To prevent intractable computation load, the proposed VIO is sliding-window and keyframe-based.
We propose to probabilistically marginalize the control points to keep the constant number of keyframes in the sliding window. Furthermore, the line exposure time difference (line delay) of the rolling shutter camera can be online calibrated in our continuous-time VIO.
To extensively examine the performance of our continuous-time VIO, experiments are conducted on publicly-available WHU-RSVI, TUM-RSVI, and SenseTime-RSVI rolling shutter datasets. The results demonstrate the proposed continuous-time VIO significantly outperforms the existing state-of-the-art VIO methods. 
%
The codebase of this paper will also be open-sourced at \url{https://github.com/APRIL-ZJU/Ctrl-VIO}.
%
%
\end{abstract}

\section{INTRODUCTION}
A wide range of sensors can be applied for accurate 6-DoF motion estimation, among which camera has become a good choice due to its low cost, light weight and intuitive perception of the appearance information. While visual odometry (VO) is able to estimate the up-to-scale camera poses, it is prone to failure when facing challenges from deficient texture, light variations and violent motion, etc.  By additionally fusing Inertial Measurement Unit (IMU) data, visual-inertial odometry (VIO) can estimate camera poses with absolute scale and becomes more robust against the aforementioned challenges compared to VO. The superiority of VIO has been shown in many existing works
~\cite{qin2018vins, geneva2020openvins, von2018direct, mourikis2007multi,mo2022continuous}.  

Most cameras can be divided into global shutter (GS) and rolling shutter (RS). Compared to GS cameras, the RS cameras are usually at a lower cost, thus they have been widely applied in consumer-grade electronics such as smartphones. However, unlike the GS cameras exposing all pixels simultaneously, the pixels of the RS cameras are exposed row by row and the timestamps of two consecutive image rows differ by a constant value termed as line delay~\cite{oth2013rolling}. Hence, significant motion distortion can be introduced in the imaging process of the RS cameras, which is named the RS effect. Some existing VIO methods just ignore the RS effect, which can severely hurt the accuracy and robustness of the estimator, especially in fast motion scenarios.

\begin{figure}[t]
    \centering
    \includegraphics[width=0.9\linewidth]{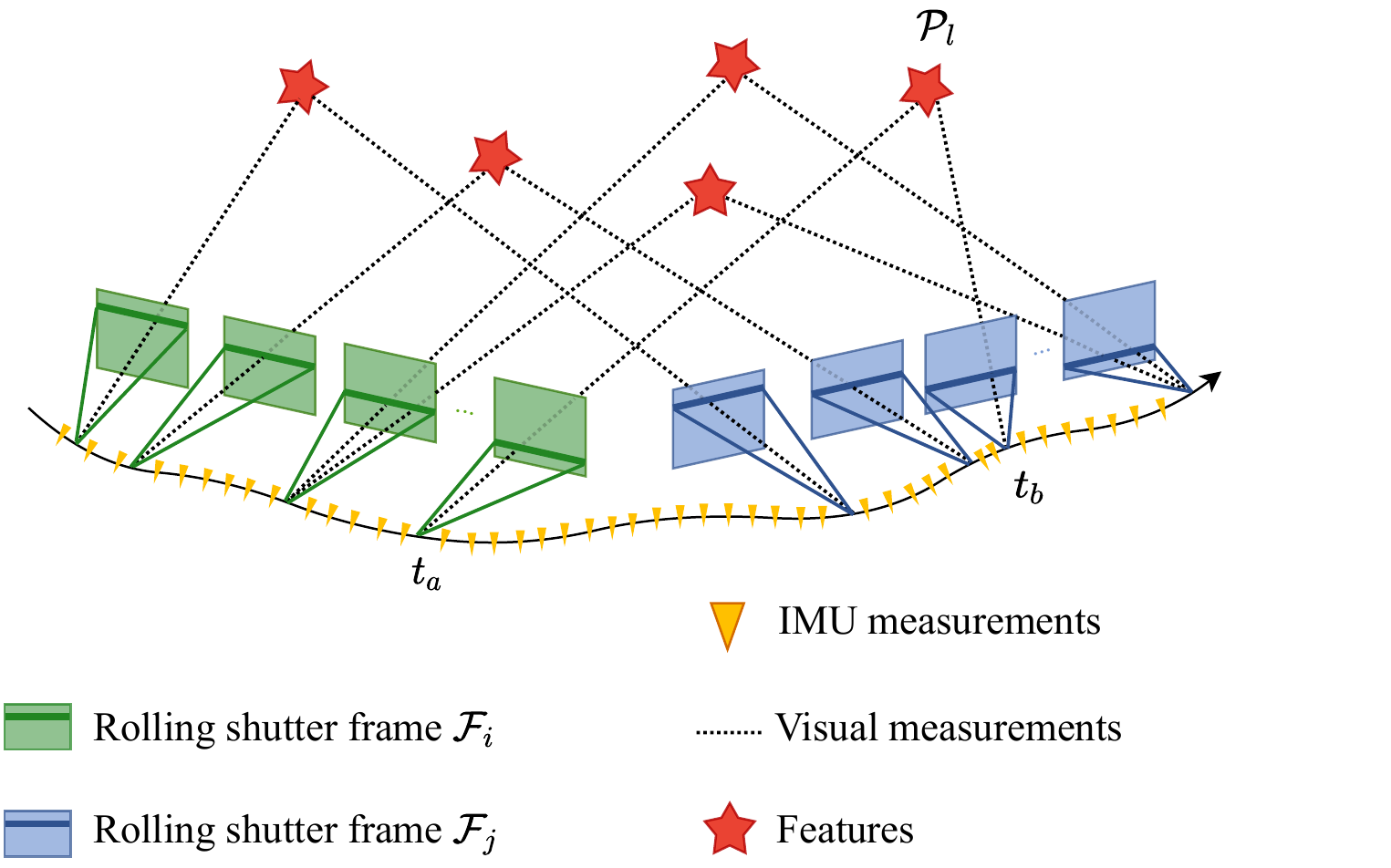}
    \captionsetup{font={small}}
    \caption{A segment of the continuous-time trajectory of IMU in the sliding window constrained by visual measurements and IMU measurements. 
    The extrinsic transformation between the camera and IMU is omitted for simplicity. 
    $\mathcal{F}_i$ and $\mathcal{F}_j$ are two frames in the sliding window. A landmark $\mathcal{P}_l$, is first observed in frame $\mathcal{F}_i$ at time $t_a$ and later observed in frame $\mathcal{F}_j$ at time $t_b$ (see Eq.~\eqref{eq:r_c}).}
    \label{fig:sw-opt}
    \vspace{-2em}
\end{figure}

In fact, a GS image corresponds to only one camera pose, while every row of a RS image corresponds to one camera pose, inevitably leading to a sharp increase in the dimension of the states to be estimated. Therefore, it is computationally intractable to merely estimate the poses of different rows of RS images. A common way to cope with this problem is to introduce a constant velocity model, assuming the camera moves at a constant speed between two keyframes~\cite{klein2009parallel, schubert2018direct, schubert2019rolling}. Another way is to parameterize the continuous-time trajectory by B-splines~\cite{lovegrove2013spline, patron2015spline, kim2016direct, kerl2015dense}, which is a more elegant way compared to the previous method. The constant velocity assumption does not hold in the case of large accelerations, while the continuous-time splines can instead parameterize trajectory with motions at high complexity. 

To deal with the RS effect, line delay needs to be known for calculating the exact timestamp of each image row. However, in some electronic devices, line delay can not be accessed by consumers and may change with exposure configurations of the camera. To calibrate the line delay of a RS camera, there are hardware-based methods~\cite{meingast2005geometric} and calibration-pattern-based methods~\cite{oth2013rolling, huai2021continuous}, while online self-calibration methods~\cite{li2014high, yang2022online} without using specific calibration patterns might be more convenient.

Inspired by the work above, we also utilize B-splines to parameterize the camera trajectory for elegantly handling the RS effect. In previous works, such continuous-time representation has been employed for RS visual odometry~\cite{lovegrove2013spline, patron2015spline, kim2016direct, kerl2015dense} and offline calibration between RS cameras and IMU~\cite{oth2013rolling, lovegrove2013spline, patron2015spline, huai2021continuous}, none of which are complete implementation of visual-inertial odometry for RS cameras, while this paper proposes continuous-time visual-inertial odometry termed as Ctrl-VIO for RS cameras with line delay online calibration, which demonstrates high accuracy even under severe RS effects. To the best of our knowledge, this is the first continuous-time visual-inertial odometry for RS cameras with line delay online calibration. Briefly, our contributions are as follows: 
\begin{itemize}
	\item We propose a keyframe-based sliding-window visual-inertial odometry (Ctrl-VIO) for RS cameras, using continuous-time trajectory parameterized by B-splines to handle the RS effect. A dedicated marginalization strategy is proposed for the continuous-time sliding-window estimator.
	\item Online calibration for line delay is naturally supported in our system, making the best of the fact that B-splines provide closed-form analytical derivatives w.r.t temporal variables~\cite{sommer2020efficient, furgale2013unified, cioffi2022continuous}.
	\item We test our method on both synthetic and real-world data, demonstrating that our system performs much better with high accuracy and robustness on RS data than other state-of-the-art RS methods and GS methods which ignore the RS effect. Also, online-calibrated line delay can converge quickly and accurately. 
\end{itemize} 

This introduction section will be followed by the related works. Afterward, the methodology of the proposed Ctrl-VIO is elaborated in Sec.~\ref{sec:method}. Experimental results are demonstrated and discussed in Sec.~\ref{sec:exp}. Finally, we conclude the paper and look into the future in the last section.
\section{RELATED WORK}

\subsection{Rolling Shutter VSLAM/VISLAM}
Klein et al.~\cite{klein2009parallel} are the first to deal with the RS effect in VSLAM. Aiming at the defects of the iPhone imaging device such as the RS effect, they make a series of modifications to PTAM~\cite{klein2007parallel} to run the algorithm on the iPhone. They first correct the observations on RS images based on the constant velocity assumption and then feed these corrected observations into bundle adjustment. David et al. extend DSO~\cite{engel2017direct} to DSORS~\cite{schubert2018direct}, additionally estimating the velocity at each keyframe and imposing a constant velocity prior to the photometric optimization, which prevents the ambiguity of RS images from crashing the system. The system outperforms state-of-the-art GS-method VO on challenging RS sequences. Combined with~\cite{von2018direct}, DSORS is further extended to ~\cite{schubert2019rolling} by fusing inertial measurements, increasing the accuracy and stability of VO and also rendering the scale observable. 

Based on the constant velocity model, which is an approximate and low-dimensional representation of the camera trajectory, the RS effect can be partially addressed. However, such representation can lead to a loss of accuracy when the camera undergoes significant accelerations.~\cite{lovegrove2013spline, patron2015spline} propose a continuous-time representation to elegantly parameterize the camera trajectory with B-splines, avoiding lowering the dimension of the motion. They use this formulation to incorporate measurements from RS cameras and IMU, developing a VO system and a visual-inertial calibration system for RS cameras. Using the continuous-time representation, Kim et al.~\cite{kim2016direct} extend LSD-SLAM~\cite{engel2014lsd} to a RS version, turning the pose estimation problem into solving control points of B-splines in photometric bundle adjustment optimization. Similarly,~\cite{kerl2015dense} proposes a dense spline-based continuous-time tracking and mapping method for RS RGB-D cameras, modeling RS in both the RGB and depth image by a photometric error term and a geometric error term respectively.

\subsection{Line Delay Calibration}
Geyer et al.~\cite{meingast2005geometric} propose a line delay calibration method aided by a LED, where the RS camera is exposed with a LED flickering at high frequency and the resulting images include light and dark lines, the spatial frequency of which is related to the line delay and the known LED frequency. However, it requires expensive hardware and tends to be imprecision. Oth et al.~\cite{oth2013rolling} present an offline calibration procedure to determine the line delay with higher accuracy that only requires videos of a known calibration pattern, in which the continuous-time optimization is adopted.~\cite{huai2021continuous} also proposes a continuous-time spline-based approach for spatiotemporal calibration of the system composed of a RS camera and an IMU using a specific calibration target, where line delay is also accurately calibrated as a byproduct. However, offline calibration with the need for calibration targets might be tedious in some cases and online calibration can be a more friendly way, especially when the calibrated parameter changes every time the system starts. Built upon the discrete-time VIO, MSCKF~\cite{mourikis2007multi}, Li and Mourikis~\cite{li2014high} propose a high-precision pose estimation algorithm for systems equipped with low-cost inertial sensors and RS cameras. The key characteristic of the proposed method is that parameters of the RS camera and IMU can be calibrated online including the line delay. Yang et al.~\cite{yang2022online} comprehensively investigate the online self-calibration of visual-inertial state estimation in the MSCKF framework, which also supports line delay calibration of RS cameras.

Compared to the aforementioned work, our method not only utilizes the continuous-time
B-spline trajectory to tightly fuse visual and inertial measurements in the odometry system, but also elegantly handles the RS effect. We are among the first to propose a complete continuous-time sliding-window VIO for RS cameras, implemented as a keyframe-based estimator with probabilistic sliding-window optimization. Carefully-designed marginalization accommodating continuous-time optimization is proposed for the sliding-window estimator.

\section{METHODOLOGY}
\label{sec:method}

We first introduce the convention in this paper. We denote the 6-DoF rigid transformation by ${}^A_B\bT \in SE(3) \in \mathbb{R}^{4\times4}$, which transforms the point ${}^B\bp$ in the frame $\{B\}$ into the frame $\{A\}$. ${}^A_B\bT=\begin{bmatrix} {}^A_B\bR & {}^A\bp_B \\ \mathbf{0} & 1 \end{bmatrix}$ 
consists of rotation ${}^A_B\bR \in SO(3)$ and translation ${}^A\bp_B \in \mathbb{R}^{3}$. Exp$(\cdot)$ maps Lie Algebra to Lie Group and Log$(\cdot)$ is its inverse operation. $(\cdot)_{\wedge}$ maps a 3D vector to the corresponding skew-symmetric matrix, while $(\cdot)_{\vee}$ is its inverse operation~\cite{barfoot2017state}.

Pixels at different rows of a RS image are exposed at different time instants. The line delay between two adjacent rows is denoted by $t_{r}$, and the timestamp of a RS image is deemed as the time instant of its first row. Hence, for a RS image with timestamp $t_i$ which has $h$ rows, the timestamps of each row are between $[t_i, t_i+ h \cdot t_r]$.
\subsection{Continous-Time Trajectory Representation}
\label{sec:continuous_time_trajectory}
In this paper, we adopt a split representation of continuous-time trajectory as~\cite{haarbach2018survey}, using two uniform cumulative B-splines to separately parameterize the 3D rotation and the 3D translation, that is 
\begin{align}
\label{eq:split_cttraj_representation}
    & \bR(u)=\bR_{i} \cdot \prod_{j=1}^{k-1} \operatorname{Exp}\left(\zeta_{j}(u) \cdot \text{Log}\left(\mathbf{R}_{i+j-1}^{-1} \mathbf{R}_{i+j}\right)\right) \\
    & \bp(u)=\bp_{i}+\sum_{j=1}^{k-1} \zeta_{j}(u) \cdot \left(\bp_{i+j} - \bp_{i+j-1} \right)  \\
    & u(t) = \frac{t-t_{0}}{\Delta t} - i, t \in [t_{i}, t_{i+1})
\end{align}
where $\bR_x$ and $\bp_x$ denote \textbf{C}ontrol \textbf{P}oints (\textbf{CP}). The knots  $t_{0}, t_{1}, t_{2}, \ldots \ $ of B-splines are uniformly distributed by the time interval $\Delta t$, thus cumulative matrix $\widetilde{\mathbf{M}}^{(k)} \in \mathbb{R}^{k \times k}$ is constant. $\zeta_{j}(u)$ is the $j$th row of $\widetilde{\mathbf{M}}^{(k)}\begin{bmatrix} 1 & u & \ldots & u^{k-1} \end{bmatrix}^\top$~\cite{sommer2020efficient}. Cubic B-Spline is adopted in this paper, which implies $k = 4$. The continuous-time trajectory of IMU in world frame $\{W\}$ can be denoted as
\begin{align}
{}_{B}^{W} \bT(t)=\left[\begin{array}{cc}{}^{W}_{B}\bR(t) & {}^{W}\bp_{B}(t) \\ 
\boldsymbol{0} & 1\end{array}\right]\text{.}
\end{align}
With the known pre-calibrated extrinsic ${}^{B}_{C}\bT$ between IMU and camera, the camera trajectory can be calculated as
\begin{align}
    {}^W_C\bT(t)={{}^W_B\bT(t)} {_{C}^{B}\bT}\text{.}
\end{align}
In this paper, we use $\bP_x$ to denote the compound control point including both the rotational CP $\bR_x$ and the positional CP $\bp_x$.
For simplicity, the four compound control points at time $t_x$ consisting of $\{\bP_{x1}, \bP_{x2}, \bP_{x3}, \bP_{x4}\}$ are also represented by a set $\mathbf{\Phi}(t_x)$, and the control points in the time interval $\left[t_x, t_{x+1}\right]$ are denoted as $\mathbf{\Phi}(t_x, t_{x+1})$. 

\begin{figure*}[t]
    \centering
    \includegraphics[width=0.8\textwidth]{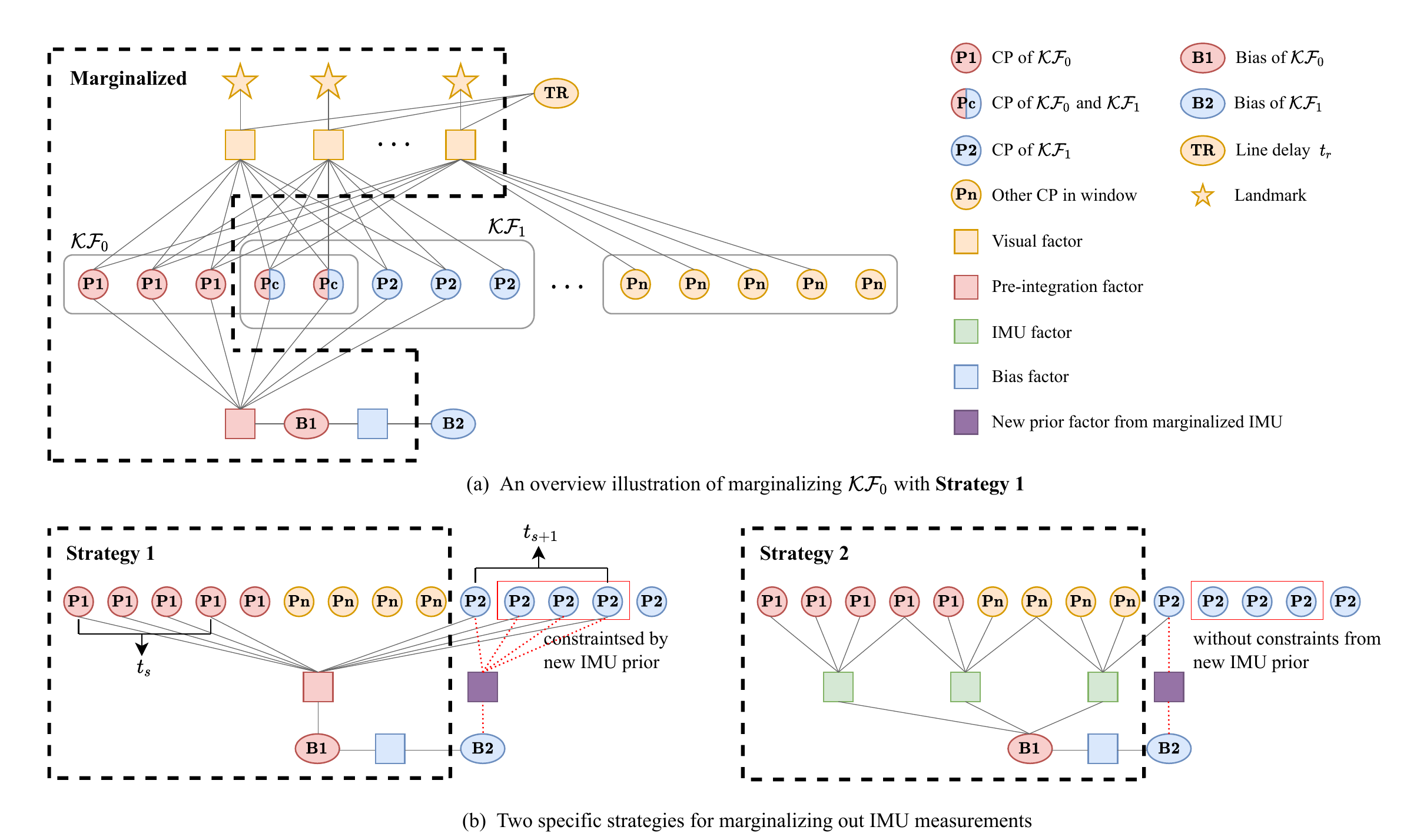}
    \captionsetup{font={small}}
    \caption{
    (a) An illustration of marginalizing out the oldest keyframe ($\mathcal{KF}_0$) including visual measurements of $\mathcal{KF}_0$ and IMU measurements in $\left[t_s, t_{s+1}\right)$. The control points (CP) to be marginalized out are the CPs of $\mathcal{KF}_0$ (red circle) expect for the CPs both involved in the second oldest keyframe $\mathcal{KF}_1$ (bicolor circle), since they will be further optimized in the next window.
    (b) Two different strategies for marginalizing IMU measurements while IMU-irrelevant factors are omitted for clearness. We take the case that there are no coupled CPs between $\mathcal{KF}_0$ and $\mathcal{KF}_1$ as example, which clearly illustrates the CPs involved with the new IMU priors resulting from two marginalization strategies are different as detailed in Sec.~\ref{sec:marginalization}.
    }
    \vspace{-1em}
    \label{fig:marg}
\end{figure*}

\subsection{Visual Factor with Line Delay}


We detect Shi-Tomasi corners~\cite{shi1994good} in RS images and use the KLT sparse optical flow~\cite{lucas1981iterative} to track the detected corners between consecutive images. Since simple homographic and fundamental checking are prone to be affected by the RS effect~\cite{hedborg2012rolling}, we adopt cross-checking to remove outliers in feature tracking. Specifically, features are firstly tracked forward in time and then tracked reversely. Only the tracked features located in the vicinity of their original positions within a certain threshold are considered inliers. 

As shown in Fig~\ref{fig:sw-opt}, a landmark $\mathcal{P}_l$, first observed in frame $\mathcal{F}_i$ with timestamp $t_i$ and later observed in frame $\mathcal{F}_j$ with timestamp $t_j$, is parameterized in inverse depth representation.
The corresponding visual factor based on reprojection error is defined as:
\begin{align} 
\label{eq:r_c}
\br_{c} &= 
\begin{bmatrix} \be_1^{\top} \\ \be_2^{\top} \end{bmatrix} \left(
\frac{\hat{\bp}_{b}}{\be_{3}^{\top} \hat{\bp}_{b}} - \pi_c\left( \begin{bmatrix} u_l^{c_b} \\ v_l^{c_b} \end{bmatrix}\right) \right) \\
\hat{\bp}_{b} &= \left({}^W_C\bT(t_b)\right)^{\top}\ {}^W_C\bT(t_a) \ \frac{1}{\lambda_{l}}
\pi_c( \begin{bmatrix} u_l^{c_a} \\ v_l^{c_a} \end{bmatrix}) \nonumber\\
t_a &= t_i + v_{l}^{c_a} \cdot t_r,\  t_b = t_j + v_{l}^{c_b} \cdot t_r \nonumber 
\end{align}
where $\be_{i}$ is a $3\times1$ vector with its $i$-th element to be 1 and the others to be 0. $\pi_c(\cdot)$ denotes the back projection function which transforms a pixel to the normalized image plane. $\begin{bmatrix} u_l & v_l \end{bmatrix}^\top$ is 2D raw observation of landmark $\mathcal{P}_l$ whose inverse depth $\lambda_l$ is defined in its first observed frame. The exact timestamps of the observations in $\mathcal{F}_i$ and $\mathcal{F}_j$ are $t_a$ and $t_b$, respectively. Note that visual landmarks are triangulated by the specific row poses of RS keyframes in the sliding window, as depicted in Fig~\ref{fig:sw-opt}, so as to initialize inverse depths of the landmarks.

From Eq.(\ref{eq:r_c}), we can find $\br_{c}$ is relevant to both $t_{a}$ and $t_{b}$, thus the Jacobian of $\br_{c}$ w.r.t $t_{r}$ is as follows.
\begin{align}
    \label{eq:rc_tr}
    \frac{\partial \br_{c}}{\partial t_{r}} = 
    \frac{\partial \br_{c}}{\partial t_{a}}
    \frac{\partial t_{a}}{\partial t_{r}}
    +
    \frac{\partial \br_{c}}{\partial t_{b}}
    \frac{\partial t_{b}}{\partial t_{r}}.
\end{align}
where $\partial \br_{c} / \partial t_{a}$ and $\partial \br_{c} / \partial t_{b}$ contain the time derivatives of splines which fortunately have analytic forms~\cite{sommer2020efficient}.
With a slight abuse of notion, we use the followings to denote the time derivative of continuous-time trajectory (refer to~\cite{lovegrove2013spline, sommer2020efficient} for the exact expressions):
\begin{align}
    \label{eq:R_t_deri}
    & \frac{\partial {}^{W}_{B}\bR(t)}{\partial t} = {}^{W}_{B}\dot{\bR}(t),
    \quad
    \frac{\partial {}^{W}\bp_{B}(t)}{\partial t} = {}^{W}\bv_{B}(t)
\end{align}
Based on Eq.(\ref{eq:r_c}, \ref{eq:rc_tr}, \ref{eq:R_t_deri}), we can derive the Jacobian of $\br_{c}$ w.r.t $t_{r}$ straightforwardly:
\begin{align}
    \frac{\partial \br_{c}}{\partial t_{r}} = 
    & \begin{bmatrix} \be_1^{\top} \\ \be_2^{\top} \end{bmatrix}
    \Big(
    v_{l}^{c_{b}} \cdot 
    {}^B_C\bR^{\top}
    {}^{W}_{B}\dot{\bR}^{\top}(t_{b})
    {}^{W}_{B}\bR(t_{a}) 
    \hat{\bp}_{m} \nonumber \\
    & +
    v_{l}^{c_{a}} \cdot
    {}^B_C\bR^{\top}
    {}^{W}_{B}\bR^{\top}(t_{b})
    {}^{W}_{B}\dot{\bR}(t_{a})
    \hat{\bp}_{m} \nonumber \\
    & + v_{l}^{c_{b}} \cdot
    {}^B_C\bR^{\top} 
    {}^W_B\dot{\bR}^{\top}(t_b)
    \left(
    {}^{W}\bp_{B}(t_{a}) - {}^{W}\bp_{B}(t_{b})
    \right) \nonumber \\
    &   + {}^B_C\bR^{\top} 
    {}^W_B\bR^{\top}(t_b)
    \left(
    v_{l}^{c_{a}} \cdot {}^{W}\bv_{B}(t_{a}) - v_{l}^{c_{b}} \cdot {}^{W}\bv_{B}(t_{b})
    \right) 
    \Big) \nonumber \\
    \hat{\bp}_{m} = 
    & {}^B_C\bR \frac{1}{\lambda_{l}}
    \pi_c( \begin{bmatrix} u_l^{c_a} \\ v_l^{c_a} \end{bmatrix}) + {}^{B}\bp_{C}
\end{align}
With the Jacobian of $\br_{c}$ w.r.t $t_{r}$, we can optimize the line delay $t_{r}$ in the factor graph optimization. 

\subsection{Inertial Factors}\label{sec:inertialfactors}
The time derivatives of the continuous-time trajectory represented by splines lead to the angular velocity and linear acceleration, which can be easily computed~\cite{sommer2020efficient}. Thus we directly utilize raw IMU measurements, consisting of angular velocity and linear acceleration, to formulate IMU factors at each IMU timestamp. Considering the IMU measurements within two consecutive frames $\mathcal{F}_k$ and $\mathcal{F}_{k+1}$ in the window, the IMU factor and the bias factor based on bias random walk are defined as:
\begin{align}
    \br_{i} &=
    \begin{bmatrix}
    {}^B{\boldsymbol{\omega}}(t)-{}^B{\boldsymbol{\omega}_m} + \bb_{g_{k}} \\ 
    {}_B^W \bR^{\top}(t)\left({}^{W}\ba(t)+{ }^{W} \bg\right) - {}^B{\ba_m} + \bb_{a_{k}} 
    \end{bmatrix} \label{eq:imu_factor}\\
    \br_{b} &= 
    \begin{bmatrix}
    \bb_{g_{k+1}} - \bb_{g_{k}} \\
    \bb_{a_{k+1}} - \bb_{a_{k}}
    \end{bmatrix}
    \label{eq:bias_factor}
\end{align}
where $ {}^B{\boldsymbol{\omega}}(t) = \left({}^W_B\bR^{\top}(t){}^W_B\dot{\bR}(t)\right)_{\vee}, {}^W\ba(t) = {}^W\ddot{\bp}_{B}(t)$ ~\cite{sommer2020efficient} are the angular velocity in $\{B\}$ frame and linear acceleration in $\{W\}$ frame, while ${}^B{\boldsymbol{\omega}_m}, {}^B{\ba_m}$ are the raw measurements of angular velocity and linear acceleration at time $t$, respectively. ${}^{W}\bg = \begin{bmatrix}0 & 0 & 9.8\end{bmatrix}^{\top}$ is the gravity vector in the world frame. Note that, the IMU measurements in $\left[t_k,t_{k+1}\right)$ share the same gyroscope bias $\bb_{g_k}$ and accelerometer bias $\bb_{a_{k}}$.

IMU factors formulated with raw IMU measurements in the sliding-window optimization is displayed in Fig.~\ref{fig:sw-opt}. Apart from these IMU factors, we also have an alternative option to utilize IMU pre-integration~\cite{forster2015imu,qin2018vins} factor for effective marginalization purpose (see Sec.\ref{sec:marginalization}).
With $\boldsymbol{\alpha}_{B_{k}B_{k+1}}$, $\bR_{B_{k}B_{k+1}}$ and $\boldsymbol{\beta}_{B_{k}B_{k+1}}$ denoting the IMU pre-integration measurements based on IMU raw measurements during two keyframed images $\left[t_k,t_{k+1}\right)$, the IMU pre-integration factor constraining the relative pose between two sequential keyframe images is formulated as~\cite{forster2015imu,qin2018vins}:
\resizebox{\linewidth}{!}{
\begin{minipage}{\linewidth}
\begin{align}
    & \br_{t} = \nonumber \\
    & \begin{bmatrix}
    {}^{W}_{B_k}\bR^{\top} 
    \left(
    {}^W\bp_{B_{k+1}}  - 
    {}^{W}\bp_{B_{k}}  - {}^{W}\bv_{B_{k}}  \Delta t +
    \frac{1}{2}\bg^{W} \Delta t ^{2}
    \right) - \boldsymbol{\alpha}_{B_{k}B_{k+1}}
    \\
    \text{Log}
    \left(
    \bR^{\top}_{B_{k}B_{k+1}}
    {}^{W}_{B_{k}}\bR^{\top}
    {}^{W}_{B_{k+1}}\bR
    \right)
    \\
    {}^{W}_{B_{k}}\bR^{\top}
    \left(
    {}^{W}\bv_{B_{k+1}} -
    {}^{W}\bv_{B_{k}} +
    {}^{W}\bg \Delta t
    \right) - \boldsymbol{\beta}_{B_{k}B_{k+1}}
    \end{bmatrix} 
    \label{eq:preinte_factor}
\end{align}   
\end{minipage}
}
where ${}^{W}_{B_k}\bR = {}^{W}_{B}\bR(t_{k})$ and ${}^{W}\bv_{B_{k}} = {}^{W}\bv_{B}(t_{k})$.



\subsection{Sliding-Window Optimization}
We adopt the same keyframe selection strategy as ~\cite{qin2018vins}, and perform the discrete-time VIO initialization method~\cite{qin2018vins} to boost our continuous-time VIO.
Despite the RS effect, the discrete-time VIO initialization procedure can provide relatively good initial values of gyroscope bias, gravity, landmarks and initial IMU poses, which are used as initial values in the continuous-time sliding-window optimization when the system starts. After initialization, new control points with constant time interval are added and initialized by the predicted poses from IMU measurements. 

As shown in Fig.~\ref{fig:sw-opt}, the RS visual observations and IMU raw measurements are used in the sliding-window optimization jointly. Assuming the $N$ RS image frames in the current sliding window are with timestamps $\{t_{s}, t_{s+1}, \cdots,  t_{s+N-1} \}$, and the height of RS images is $H$, then the control points $\mathbf{\Phi}(t_{s}, t_{s+N-1} +  H\cdot t_r)$ which parameterize the trajectory in the window are going to be optimized by solving the following problem:  
\resizebox{\linewidth}{!}{
\begin{minipage}{\linewidth}
\begin{align}
\label{eq:cost_func}
    & \arg \min _{\mathcal{X}} 
    \left\{
    \sum \left\|\br_{c}\right\|_{\Sigma_{c}}^{2}+
    \sum \left\|\br_{i}\right\|_{\Sigma_{i}}^{2}+
    \sum \left\|\br_{b}\right\|_{\Sigma_{b}}^{2}+
    \left\|\br_{p}\right\|^{2}
    \right\}  \text{,}
\end{align}
\end{minipage}
}
\begin{align}
    \mathcal{X} =
    \lbrace
    &\boldsymbol{\Phi}(t_s, t_{s+N-1}+H\cdot t_r), \bb_{g_{0}} \ldots \bb_{g_{N-1}}, \nonumber
    \\ & \bb_{a_{0}} \ldots \bb_{a_{N-1}}, {\lambda}_{l_{0}} \ldots {\lambda}_{l_{M-1}}, t_{r}  \nonumber
    \rbrace
\end{align}
where $M$ is the number of landmarks in the sliding window. $\br_{p}$ is the prior factor obtained by marginalization (see Sec.\ref{sec:marginalization}) , and $\Sigma_{c}$, $\Sigma_{i}$, $\Sigma_{b}$ are the corresponding covariance matrices, respectively.



\subsection{Marginalization in Continuous-Time Sliding-Window Optimization}
\label{sec:marginalization}


In order to maintain tractable computation for the sliding-window optimization, we only keep a constant number of RS image frames in the sliding window and marginalize RS frames and relevant states strategically, which is inspired by~\cite{shen2016initialization,qin2018vins}. 
When the second latest RS image frame is a non-keyframe, we directly discard visual measurements of this frame, while its corresponding control points and IMU measurements are still retained in the window.
On the other hand, when the second latest image frame is a keyframe, we marginalize the oldest keyframe with its corresponding measurements, while reserving part of its control points involved in the remaining frames. That is, the control points that are older than the first control point of the second oldest keyframe $\bPhi_{\text{marg}} = \bPhi(t_s,t_{s+1}) - \bPhi(t_{s+1})$, the IMU biases $\{\bb_{g_s}, \bb_{a_s}\}$, and the inverse depths of the features in the oldest keyframe, are marginalized out. 

In this paper, we design the following two options to marginalize the IMU relevant measurements~\ref{sec:inertialfactors} after the sliding-window optimization:

\noindent\textbf{Strategy 1}: pre-integrate all the intermediate IMU raw measurements between the oldest keyframe and the second oldest keyframe, that is $\left[t_s, t_{s+1}\right)$, formulate one pre-integration~\cite{forster2015imu} factor (Eq.~\ref{eq:preinte_factor}) and one bias factor (Eq.~\ref{eq:bias_factor}) with $\{\bb_{g_s}, \bb_{a_s},\bb_{g_{s+1}}, \bb_{a_{s+1}}\}$ in the sliding-window optimization, then marginalize out $\{\bPhi(t_s) - \bPhi(t_{s+1}), \bb_{g_s}, \bb_{a_s}\}$.

\noindent\textbf{Strategy 2}: directly formulate certain IMU factors (Eq.~\ref{eq:imu_factor}) with the raw IMU measurements in $\left[t_s, t_{s+1}\right)$ and one bias factor (Eq.~\ref{eq:bias_factor}) with $\{\bb_{g_s}, \bb_{a_s},\bb_{g_{s+1}}, \bb_{a_{s+1}}\}$, and marginalize out $\{\bPhi_{\text{marg}}, \bb_{g_s}, \bb_{a_s}\}$.


Strategy 2 is exactly in accordance to the optimization cost function Eq.~\eqref{eq:cost_func}, while Strategy 1 requires refactoring the cost function (Eq.~\eqref{eq:cost_func}) after optimization by replacing the IMU factor cost $\sum \left\|\br_{i}\right\|_{\Sigma_{i}}^{2}$ with IMU pre-integration factor cost $\sum \left\|\br_{t}\right\|_{\Sigma_{t}}^{2}$.
Although Strategy 2 is more straightforward and intuitive, Strategy 1 could be more effective. Figure~\ref{fig:marg} illustrates the details of the marginalization and the two strategies for marginalizing IMU measurements. The main difference is that the control points of the second oldest frame $\bPhi(t_{s+1})$ are certainly involved in the pre-integration factor in Strategy 1, while rarely fully-involved in Strategy 2.
After marginalization is carried out via Schur Complement~\cite{sibley2010sliding}, we get the prior factor which constrains the remaining control points and states in the window.

\section{EXPERIMENTS} \label{sec:exp}
We evaluate the proposed method on a synthetic  dataset WHU-RSVI~\cite{cao2020whu} and two real-world datasets TUM-RSVI~\cite{schubert2019rolling}, SenseTime-RSVI~\cite{jinyu2019survey}. 
The state-of-the-art monocular visual-inertial odometry that we compared against are two RS-aware methods, RS-VI-DSO~\cite{schubert2019rolling} and RS-VINS-Mono~\cite{qin2018vins}, and several methods without RS-aware (GS methods) including OKVIS~\cite{leutenegger2015keyframe}, VINS-Mono~\cite{qin2018vins}, ORB-SLAM3~\cite{campos2021orb}, where for fairness, we disable the loop detection and pose graph optimization of VINS-Mono and ORB-SLAM3 for pure VIO evaluations. 
RS-VINS-Mono\footnote{
\url{https://github.com/HKUST-Aerial-Robotics/VINS-Mono}}, a RS version of VINS-Mono, handles the RS effect similarly to~\cite{klein2009parallel} but requires known line delay and cannot perform online calibration of line delay. To study the main contribution of our proposed Ctrl-VIO, which is mainly the backend with sliding-window continuous-time optimization, we adapt the frontend of RS-VINS-Mono with cross-checking to remove outliers, and  ensure the frontend of the adapted RS-VINS-Mono is the same as our Ctrl-VIO for fair comparison of the backend.

We evaluate Absolute Pose Error (APE) of trajectories composed of keyframe poses by the toolbox~\cite{grupp2017evo}. Every method runs six rounds on each sequence and takes the average. 
Note that in the following experiments, Ctrl-VIO adopts Strategy 1 for marginalization by default, while Ctrl-VIO-margIMU adopts Strategy 2, and this is the only difference between the two. In our experiments, the time interval of B-splines are set as 0.03s and 0.05s for the synthetic dataset and real-world datasets respectively.
The max number of features to track is $150$, and the number of image frames $N$ in the window is $11$.

\begin{figure}[t]
	\centering
	\begin{subfigure}[b]{0.6\linewidth}
		\centering
		\includegraphics[width=\linewidth]{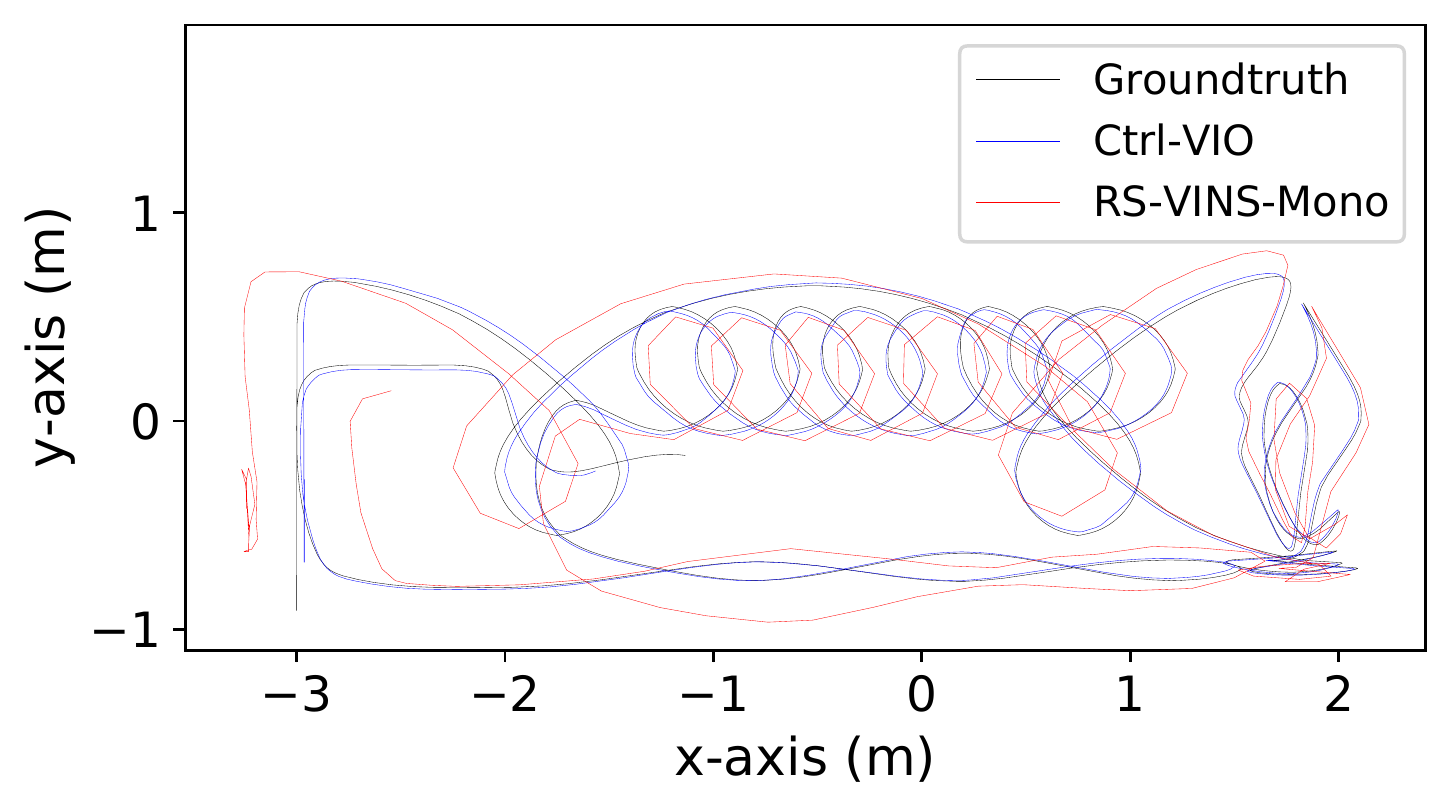}
	\end{subfigure}
	\begin{subfigure}[b]{0.6\linewidth}
		\centering
		\includegraphics[width=\linewidth]{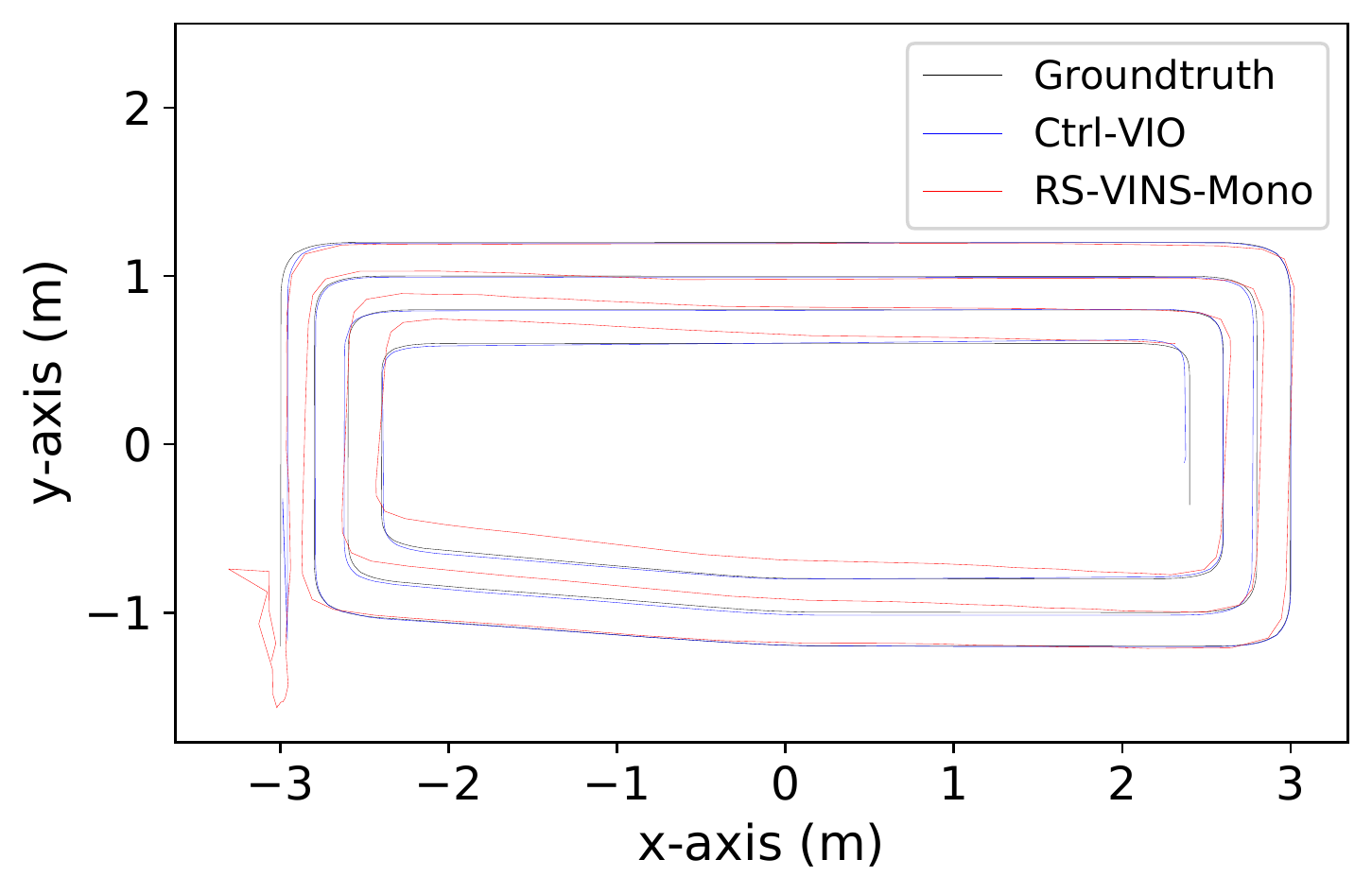}
	\end{subfigure}
	\captionsetup{font={small}}
	\caption{Estimated trajectories compared to the groundtruth in sequences fast-1 and fast-2 in WHU-RSVI dataset.}
	\vspace{-1em}
	\label{fig:traj1_traj2}
\end{figure}

\begin{table}[t]
\captionsetup{font={small}}
\caption{The RMSE(m) of APE results of different methods in WHU-RSVI dataset. The best results are marked in bold.}
\label{tab:whu_ape}
\centering
\resizebox{\linewidth}{!}{
\begin{tabular}{@{}ccccccc@{}}
\toprule
& \multicolumn{3}{c}{GS Method} & \multicolumn{3}{c}{RS Method}                                                                                                    \\ \cmidrule(r){2-4}  \cmidrule(r){5-7} 
       & OKVIS & VINS-Mono & ORB-SLAM3 & \begin{tabular}[c]{@{}c@{}}RS-\\ VINS-Mono\end{tabular} & Ctrl-VIO & \begin{tabular}[c]{@{}c@{}}Ctrl-VIO-\\ margIMU\end{tabular} \\ \midrule
\textit{fast-1} & 0.539 & 0.435 & 0.477 & 0.149 & \textbf{0.034} & 0.133 \\
\textit{medium-1} & 0.472 & 0.242 & 0.074 & 0.133 & \textbf{0.058} & 0.106 \\
\textit{slow-1} & 0.415 & 0.267 & 0.068 & 0.148 & \textbf{0.061} & 0.257 \\
\textit{fast-2} & 0.492 & 0.572 & 0.061 & 0.125 & \textbf{0.027} & 0.077 \\
\textit{medium-2} & 0.184 & 0.214 & 0.067 & 0.083 & \textbf{0.064} & 0.092 \\
\textit{slow-2} & 0.168 & 0.074 & \textbf{0.031} & 0.085 & 0.068 & 0.084 \\ \bottomrule
\end{tabular}
}
\vspace{-1em}
\end{table}

\begin{table}[t]
\captionsetup{font={small}}
\caption{ The mean and standard deviation of the calibrated line delay by Ctrl-VIO on WHU-RSVI dataset. The ground truth of line delay is 69.44 us.}
\label{tab:whu_linedelay}
\centering
\resizebox{0.6\linewidth}{!}{
\begin{tabular}{@{}ccc@{}}
\toprule
&  Calibrated line delay (us) & Error (us) \\   \midrule
\textit{fast-1}   & 69.46 $\pm$ 0.55 & 0.02  \\
\textit{medium-1} & 71.50 $\pm$ 0.31 & 2.06  \\
\textit{slow-1}   & 70.81 $\pm$ 0.15 & 1.37  \\
\textit{fast-2}   & 67.89 $\pm$ 0.41 & -1.55 \\
\textit{medium-2} & 72.06 $\pm$ 0.39 & 2.62  \\
\textit{slow-2}   & 72.45 $\pm$ 0.48 & 3.01  \\ \bottomrule
\end{tabular}
}
\vspace{-1em}
\end{table}

\begin{table*}[t]
\captionsetup{font={small}}
\caption{The RMSE(m) of APE results of GS methods on GS data, and both GS and RS methods on RS data in TUM-RSVI dataset. The best results are marked in bold and the second best results are marked underlined.}
\label{tab:tum_ape}
\centering
\resizebox{\linewidth}{!}{
\begin{tabular}{@{}ccccccccccccc@{}}
\toprule
Image Type & Method & RS-Aware & \textit{seq01} & \textit{seq02} & \textit{seq03} & seq04 & \textit{seq05} & \textit{seq06} & \textit{seq07} & \textit{seq08} & \textit{seq09} & \textit{seq10} \\ \midrule
 & OKVIS & \xmark & 0.199 & 0.227 & 0.168 & 0.151 & 0.138 & 0.212 & 0.203 & 0.156 & 0.157 & 0.134 \\
 & VINS-Mono & \xmark & 0.171 & 0.129 & 0.273 & \textbf{0.092} & 0.098 & 0.064 & 0.083 & 1.310 & 0.224 & 0.130 \\
 & ORB-SLAM3 & \xmark & \textbf{0.020} & 0.020 & 0.037 & 0.162 & \textbf{0.036} & \textbf{0.017} & \textbf{0.036} & \textbf{0.025} & \textbf{0.044} & \textbf{0.107} \\
\multirow{-4}{*}{GS} & VI-DSO & \xmark & 0.038 & \textbf{0.018} & \textbf{0.027} & 0.137 & 0.060 & 0.135 & 0.061 & 0.070 & 0.128 & 0.111 \\
\midrule
 & OKVIS & \xmark & 0.895 & 0.641 & 0.435 & 0.216 & 0.301 & 0.576 & 0.246 & 0.185 & 1.002 & 1.683 \\
 & VINS-Mono & \xmark & 114.710 & 2.230 & 72.850 & 53.130 & 122.690 & 1.570 & 155.930 & 0.409 & 130.950 & 159.910 \\
 & ORB-SLAM3 & \xmark & 2.100 & - & 0.601 & 0.204 & 0.269 & 0.936 & 0.260 & 0.365 & - & - \\
 & VI-DSO & \xmark & 79.591 & 40.725 & 1.803 & 0.970 & 0.683 & 2.352 & 28.336 & 0.501 & 218.152 & 482.021 \\
 & RS-VINS-Mono & \cmark & 0.129 & 0.102 & 0.075 & 0.049 & 0.071 & 0.115 & 0.102 & \underline{0.060} & 0.146 & 0.189 \\
 & RS-VIDSO & \cmark & \underline{0.040} & \underline {0.044} & \textbf{0.028} & 0.079 & \underline{0.049} & \textbf{0.017} & \underline{0.075} &  0.168 & 0.168 &  0.246 \\
& Ctrl-VIO & \cmark & \textbf{0.027} & \textbf{0.033} & \underline {0.047} & \textbf{0.042} & \textbf{0.041} & 0.054 & \textbf{0.057} & \textbf{0.058} &  \textbf{0.062} & \textbf{0.079} \\
\multirow{-8}{*}{RS} & Ctrl-VIO-margIMU & \cmark & 0.065 & 0.049 & 0.048 & \underline{0.046} & 0.069 & \underline{0.039} & 0.083 & 0.067 & \underline{0.097} & \underline{0.128} \\ \bottomrule
\end{tabular}
}
\vspace{-1em}
\end{table*}

\begin{figure}[t]
    \centering
    \includegraphics[width=0.9\linewidth]{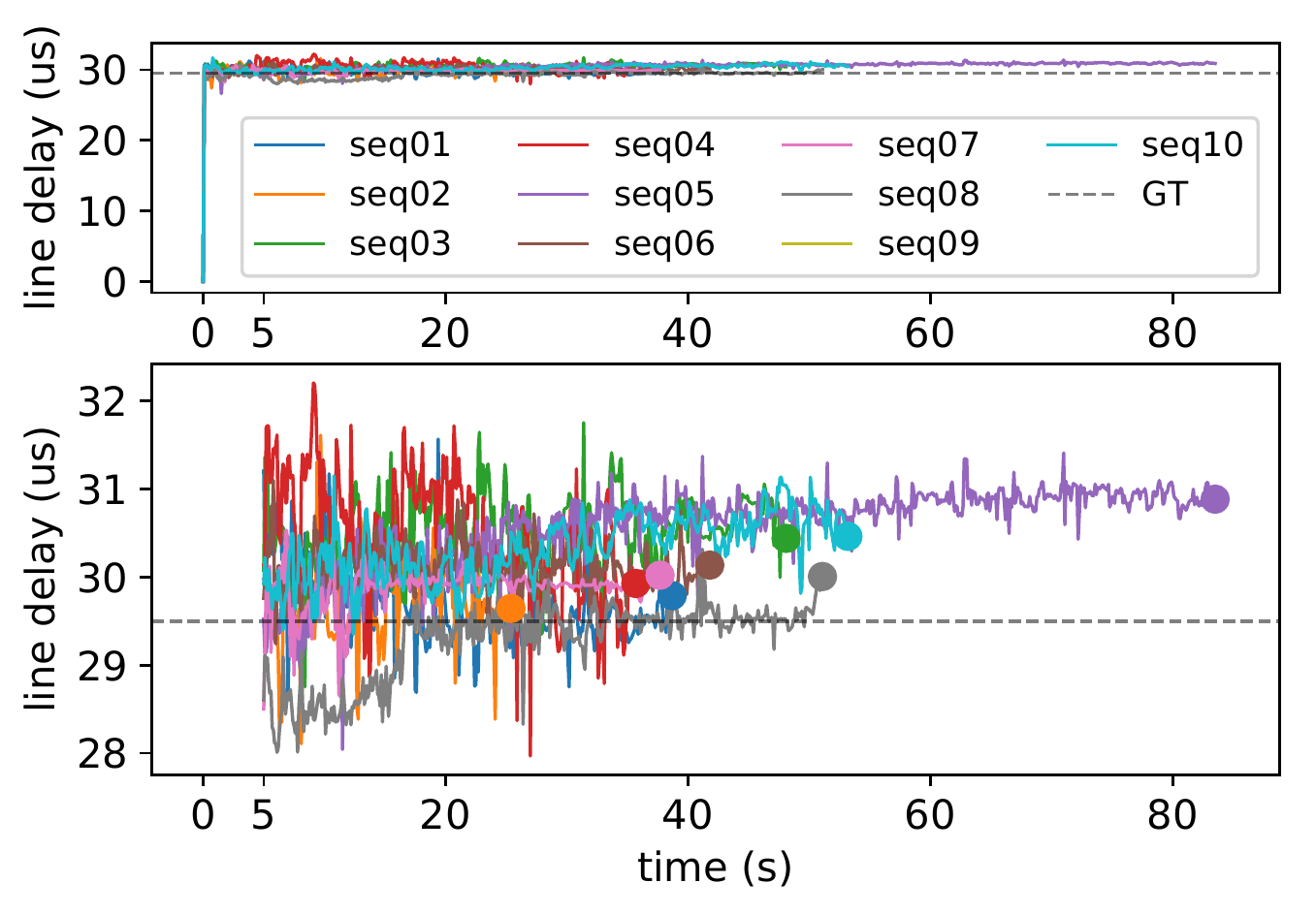}
    \captionsetup{font={small}}
    \caption{The calibrated line delay along with time in each sequence of TUM-RSVI dataset, starting from zero initial value.
    The dashed line represents the reference. Top: line delay estimations along the entire sequence. Bottom: zoom in on $[5,80]$ seconds. }
    \vspace{-1em}
    \label{fig:10bag1ld_tum}
\end{figure}

\begin{figure}[t]
    \centering
    \includegraphics[width=1.0\linewidth]{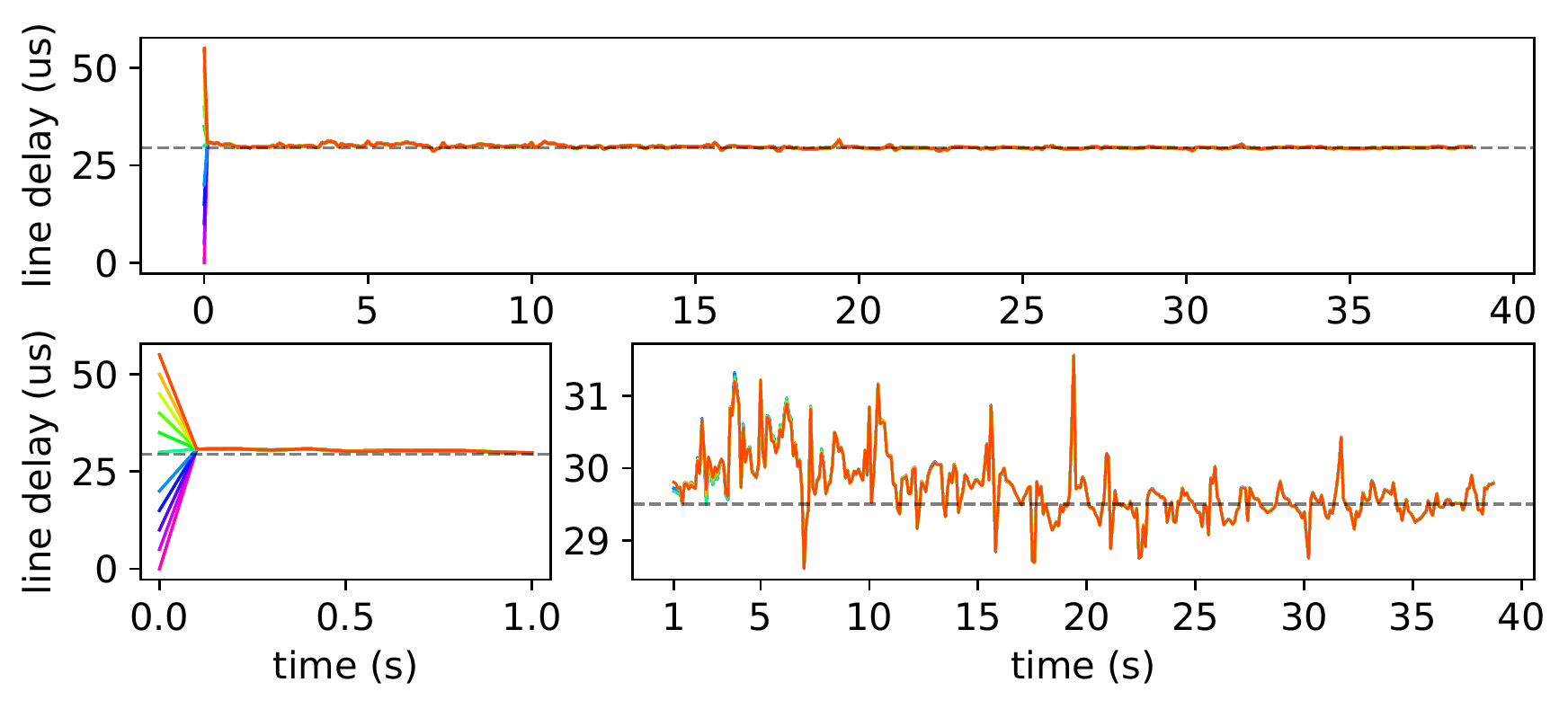}
    \captionsetup{font={small}}
    \caption{Consistent calibration results of the line dealy are achieved from various initial values, in \textit{seq01} of TUM-RSVI dataset. The dashed line represents the reference value.  Top: line delay estimations along the entire sequence. Bottom: zoom in on $[0,1]$ seconds and $[1,40]$ seconds. }
    \vspace{-1em}
    \label{fig:1bag11ld_tum}
\end{figure}

\begin{table}[t]
\captionsetup{font={small}}
\caption{The RMSE(m) of APE results of different methods in SenseTime-RSVI dataset. The best results are in bold.}
\label{tab:sensetime_ape}
\centering
\resizebox{\linewidth}{!}{
\begin{tabular}{@{}ccccccc@{}}
\toprule
& \multicolumn{3}{c}{GS Method} & \multicolumn{3}{c}{RS Method}                                                                                                    \\ \cmidrule(r){2-4}  \cmidrule(r){5-7} 
 & OKVIS & VINS-Mono & ORB-SLAM3 & \begin{tabular}[c]{@{}c@{}}RS-\\ VINS-Mono\end{tabular} & Ctrl-VIO & Ctrl-VIO-margIMU \\ \midrule
\textit{A0} & 0.072 & 0.075 & 0.089 & \textbf{0.062} & 0.068 & 0.065 \\
\textit{A1} & 0.088 & 0.161 & 0.071 & 0.044 & 0.036 & \textbf{0.034} \\
\textit{A2} & 0.068 & 0.057 & 0.107 & 0.045 & 0.047 & \textbf{0.039} \\
\textit{A3} & 0.023 & 0.024 & 0.024 & 0.016 & 0.016 & \textbf{0.015} \\
\textit{A4} & 0.147 & 0.022 & 0.028 & 0.018 & \textbf{0.017} & 0.030 \\
\textit{A5} & 0.078 & 0.050 & 0.080 & 0.059 & \textbf{0.046} & \textbf{0.046} \\
\textit{A6} & 0.064 & 0.027 & 0.045 & 0.022 & 0.019 & \textbf{0.018} \\
\textit{A7} & 0.047 & 0.020 & 0.032 & 0.012 & \textbf{0.011} & 0.016 \\ \bottomrule
\end{tabular}
}
\end{table}

\begin{figure}[t]
    \centering
    \includegraphics[width=1.0\linewidth]{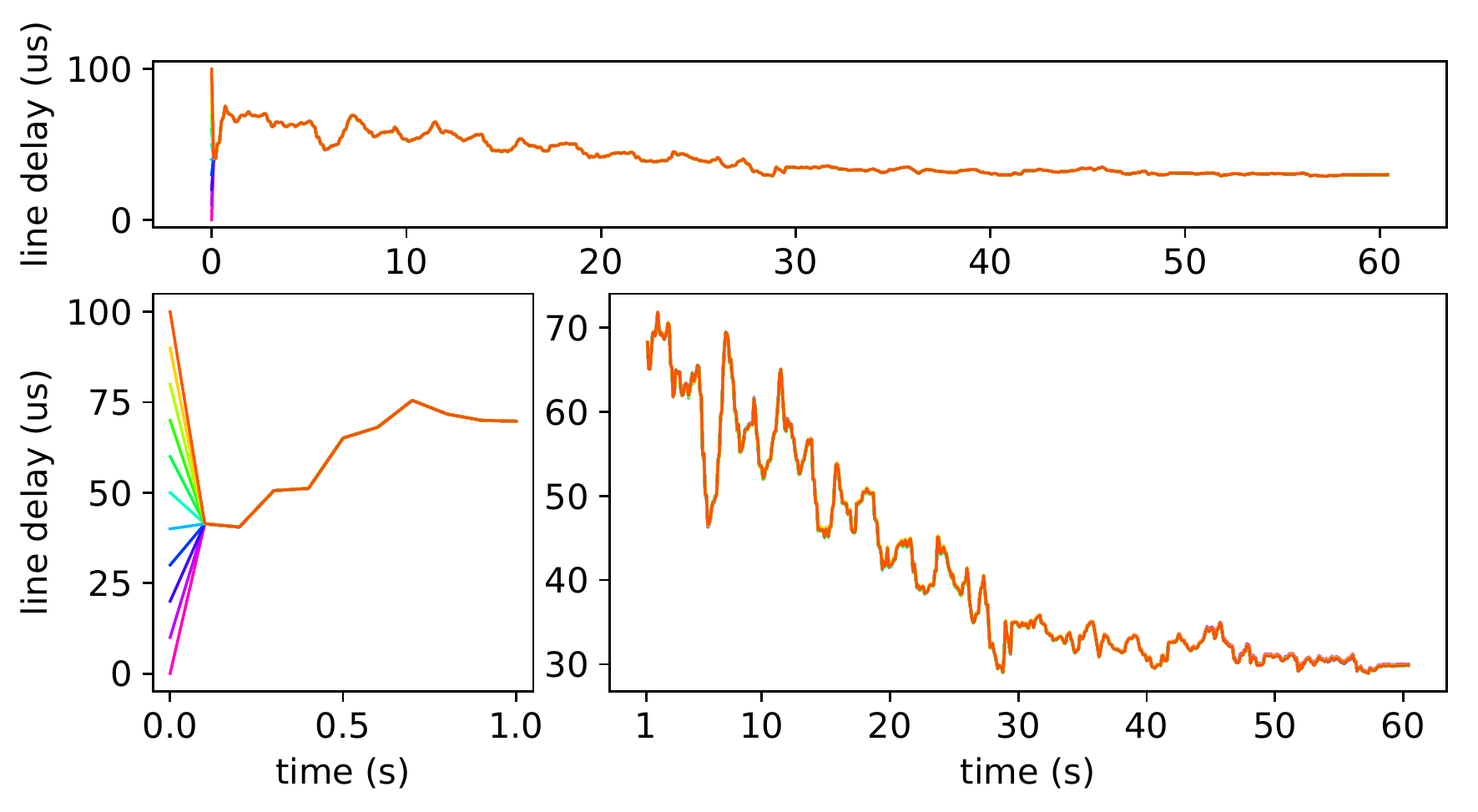}
    \captionsetup{font={small}}
    \caption{Consistent results of the line dealy are achieved with various initial values, over \textit{A6} sequence. Top: line delay estimations along the entire sequence. Bottom: zoom in on $[0,1]$ seconds and the remaining.}
    \vspace{-2em}
    \label{fig:1bag11ld_sensetime}
\end{figure}

\begin{table}[t]
\captionsetup{font={small}}
\caption{The mean and standard deviation of runtime of Ctrl-VIO and RS-VINS-Mono on \textit{seq1} of TUM-RSVI Dataset.}
\label{tab:tum_runtime}
\centering
\resizebox{1.0\linewidth}{!}{
\begin{tabular}{@{}cccc@{}}
\toprule
& Frontend (ms) & Optimization (ms)   & Marginalization (ms) \\  \midrule
RS-VINS-Mono & 27.74(16.24) & 18.96(10.29)   & 3.72(4.22) \\
Ctrl-VIO     & 27.31(16.89) & 123.01(111.96) & 11.07(12.43) \\   \bottomrule
\end{tabular}
}
\vspace{-1em}
\end{table}

\subsection{WHU-RSVI Dataset}
The WHU-RSVI Dataset~\cite{cao2020whu} synthesizes two trajectory profiles (see Fig.~\ref{fig:traj1_traj2}) at three different motion speeds (fast, medium, slow), which lead to different extent of RS effects. We name sequences with trajectory number and speed, e.g, \textit{medium-2} is a sequence of trajectory 2 with medium speed. The simulated IMU is collected at 90 Hz and the RS camera at 30 Hz, and the groundtruth of the line delay is 69.44 us. 

Firstly, we evaluate the pose estimation performance of various methods with the given ground-truth of line delay.
Table~\ref{tab:whu_ape} summarizes the RMSE of APE of different methods. With the same trajectory profile, GS methods tend to have larger APE as the motion speed increases. 
Interestingly, ORB-SLAM3 is less impacted by the RS effect, which is consistent with the experimental findings ~\cite{johansson2021evaluation}. However, images in \textit{fast-1} sequence with highly dynamic motion, especially the violent rotation, suffer from severe RS effects, resulting in the inferior performance of ORB-SLAM3. 
Ctrl-VIO achieves the best accuracy in almost all test sequences. It is also interesting that Ctrl-VIO performs better on fast motion sequences. The reason we speculate might be that highly-dynamic motions make the scale of monocular VIO systems well observable~\cite{shen2016initialization}. 
On the other hand, RS-VINS-Mono performs better than VINS-Mono but is not comparable with Ctrl-VIO, mainly because it assumes constant velocity motion to remove the RS effect, which is not a reasonable hypothesis, especially under large accelerations.  

Additional experiments are carried out to verify line delay online calibration accuracy. Starting from zero initial value, the line delay quickly converges within 1 second in our experiments. 
The calibration results are shown in Table~\ref{tab:whu_linedelay} , where we summarize the mean and standard deviation of the line delay in the last 5 seconds of each sequence. The results demonstrate that line delay is able to be calibrated by Ctrl-VIO with high accuracy and low standard deviation.

\subsection{TUM-RSVI Dataset}
The TUM-RSVI dataset~\cite{schubert2019rolling} is a handheld real-world dataset consisting of a GS camera at 20 Hz, a RS camera at 20 Hz and an IMU at 200Hz. All sensors are hardware synchronized and the line delay of the RS camera approximates 29.4737us, which is given as a reference by the dataset.
We test GS methods on both GS and RS data, and RS methods on RS data only. It is noted that RS-VINS-Mono is run with the reference line delay given by the dataset since  it does not support online calibration, while the proposed Ctrl-VIO simultaneously calibrates the line delay online from the zero initial value.

The RMSE of APE is shown in Table~\ref{tab:tum_ape}, where the RMSE of VI-DSO~\cite{von2018direct} and its RS version (named RS-VI-DSO) are directly excerpted from~\cite{schubert2019rolling}.
GS methods have plausible performance on GS data, but are deteriorated substantially on RS data, especially on \textit{seq09} and \textit{seq10}, where angular velocities are particularly large, leading to a significant RS effect. 
However, RS-aware methods which explicitly address the RS effect work well on the RS data, and can achieve comparable accuracy as GS methods on GS data. 
Ctrl-VIO outperforms RS-VINS-Mono and RS-VI-DSO on most sequences. Moreover, averagely large angular acceleration on both \textit{seq09} and \textit{seq10}, makes the constant velocity assumption used by RS-VINS-Mono and RS-VI-DSO less applicable, while Ctrl-VIO elegantly handling the RS effect with continuous-time trajectory has higher accuracy.

Fig.~\ref{fig:10bag1ld_tum} also shows the convergence of line delay over time on different sequences starting from zero initial value. The line delay parameter quickly converges to a near reference value and keeps nearly static in the remaining trajectory. In addition, we further examine the line delay calibration starting from various initial values. Fig.~\ref{fig:1bag11ld_tum} displays the line delay calibrated from initial values of 0$\sim$55 us with the interval of 5 us. The line delay can stably converge to a near reference value with various initial values.

\subsection{SenseTime-RSVI Dataset} 
The SenseTime-RSVI Dataset~\cite{jinyu2019survey} provides RS visual-inertial data collected by a mobile phone, Xiaomi Mi 8. The RS images are captured at 30Hz and IMU data collected at 400Hz. No reference value of line delay is given in this dataset. 
Fortunately, the proposed Ctrl-VIO is able to run with calibrating line delay online, which is initialized from zero. Note that, we run RS-VINS-Mono with the line delay estimated by Ctrl-VIO.
Compared to TUM-RSVI Dataset, SenseTime-RSVI Dataset is collected under more smooth and slow motion, without violent shake and significant accelerations like the former.

Table~\ref{tab:sensetime_ape} shows the RMSE of APE of different methods, where the RMSE of OKVIS and VINS-Mono are directly from~\cite{jinyu2019survey}. Ctrl-VIO gets more accurate pose estimation on almost all the sequences. It is also interesting to find that GS methods achieve satisfying accuracy mainly because of lowly-dynamic motions of the dataset. It can be easily found that RS-VINS-Mono with the line delay calibrated by Ctrl-VIO gets better performance than VINS-Mono without RS-aware.
Fig.~\ref{fig:1bag11ld_sensetime} also shows the estimated line delay over time on sequence A6, starting from different initial values ( from 0 to 100 us with an interval of 10 us). The final converged value of line delay in all trails is about 30 us which is a reasonable value and matches the conclusion in~\cite{oth2013rolling} that the line delay of RS cameras of smartphones is around 25$\sim$60 us. However, the convergence of the line delay takes longer time on this dataset, partially due to the insufficient RS effect of the lowly-dynamic motion. 

\subsection{Comparison of marginalization strategies for IMU}

On the SenseTime-RSVI dataset with lowly-dynamic motion, Ctrl-VIO and Ctrl-VIO-margIMU have very similar performances on pose estimation accuracy (see Table~\ref{tab:sensetime_ape}). However, as the pose estimation results are shown in Table~\ref{tab:whu_ape} and \ref{tab:tum_ape} over WHU-RSVI and TUM-RSVI dataset, the Ctrl-VIO with marginalization Strategy 1 outperforms Ctrl-VIO-margIMU with marginalization Strategy 2. The difference is caused by the different factor graph formulations in Strategy 1 and Strategy 2. The pre-integration factor in Strategy 1 and IMU factors in Strategy 2 involve different control points (described in Sec.~\ref{sec:marginalization}). After IMU information between the oldest keyframe and the second oldest keyframe is marginalized out, for Strategy 2, the new prior factor obtained might not constrain all control points corresponding to the timestamp of the second oldest keyframe, while Strategy 1 indeed does. Although we admit that both marginalization strategies are rational theoretically, they can have different numerical performances in the non-linear least-square optimizations.

\subsection{Runtime analysis}
We compare the runtime of Ctrl-VIO and RS-VINS-Mono on \textit{seq1} of the TUM-RSVI Dataset, and Table~\ref{tab:tum_runtime} shows the result. Both methods are executed on the desktop PC with an Intel i7-8700 CPU @ 3.2Ghz and 32GB RAM. The frequency of RS-VINS-Mono and Ctrl-VIO are 19.83Hz and 6.19Hz, respectively. And the time cost of optimization and marginalization of Ctrl-VIO have relatively high standard deviations since the sliding window duration varies. 
The number of control points to be optimized increases when the time span of the sliding window is large, and fewer control points will be optimized if the time span is small. 
For the former situation, the motion is usually gentle and we do not have to uniformly add so many control points to constrain the trajectory. 
Thus, we consider using a non-uniform B-spline 
in future work to improve the computational efficiency.




\section{Conclusions and Future Work}
This paper proposes continuous-time visual-inertial odometry dedicated to RS cameras, dubbed Ctrl-VIO. The VIO is implemented as a keyframe-based sliding-window estimator, which accommodates the continuous-time state optimization and elegantly handles the RS effects.
RS visual and IMU measurements are tightly coupled to optimize the trajectory in the sliding window. A specific marginalization method for the continuous-time sliding-window estimator is investigated, and two different ways of marginalizing IMU measurements are compared and discussed. Besides, line delay can be online calibrated in the proposed Ctrl-VIO, which is useful when the line delay is unknown. Verified by simulation and real-world experiments, our system demonstrates high accuracy on RS data, superior to the existing state-of-the-art VIO. 
Although Ctrl-VIO has good performance regarding accuracy and robustness, it cannot be run in real-time currently. The control points in the current implementation are distributed uniformly over time, which is unnecessary when the motion is slow and smooth. It is worthwhile investigating the non-uniform B-spline to parameterize the trajectory with fewer control points to improve efficiency further.

\section{Acknowledgement}
This work is partially supported by National Natural Science Foundation of China under grant NSFC 62088101.






{
\AtNextBibliography{\scriptsize}
\printbibliography
}

\end{document}